\documentclass[10pt,twocolumn,letterpaper]{article}

\usepackage[pagenumbers]{cvpr} 
\usepackage{booktabs}
\usepackage{amssymb}
\usepackage{array}
\usepackage{epsfig}
\usepackage{amsmath}
\usepackage{amssymb}
\usepackage{multicol}
\usepackage{multirow}
\usepackage{nccmath}
\usepackage{pifont}
\usepackage{adjustbox}
\usepackage{enumitem}
\usepackage{caption}
\usepackage{booktabs}
\usepackage{ragged2e}
\usepackage{comment}
\usepackage{color}
\usepackage{float}
\usepackage{tcolorbox}
\tcbuselibrary{breakable}

\usepackage{wrapfig}
\usepackage{algorithm}
\usepackage{algorithmic}

\usepackage{stfloats}

\usepackage{bbding}
\usepackage{fontawesome5}

\definecolor{cvprblue}{rgb}{0.21,0.49,0.74}
\usepackage[pagebackref,breaklinks,colorlinks,allcolors=cvprblue]{hyperref}

\definecolor{darkergreen}{RGB}{19,168,33}
\newcommand{\greencheck}{{\color{darkergreen}\ding{51}}}
\newcommand{\redcross}{{\color{red}\ding{55}}}

\newcommand\blfootnote[1]{%
  \begingroup
  \renewcommand\thefootnote{}\footnote{#1}%
  \addtocounter{footnote}{-1}%
  \endgroup
}

\def\paperID{9957} 
\def\confName{CVPR}
\def\confYear{2026}

\title{MarketGen: A Scalable Simulation Platform with Auto-Generated Embodied Supermarket Environments}

\author{
    Xu Hu\textsuperscript{1,7} \quad 
    Yiyang Feng\textsuperscript{2,7} \quad 
    Junran Peng\textsuperscript{2,7}\textsuperscript{$\dagger$\faEnvelope[regular]} \quad 
    Jiawei He\textsuperscript{5} \quad 
    Liyi Chen\textsuperscript{1} \quad \\
    Wei Sui\textsuperscript{8} \quad
    Chuanchen Luo\textsuperscript{6,7} \quad 
    Xucheng Yin\textsuperscript{2} \quad
    Qing Li\textsuperscript{1}\textsuperscript{\faEnvelope[regular]} \quad
    Zhaoxiang Zhang\textsuperscript{3,4}\textsuperscript{\faEnvelope[regular]} \\
    \small \textsuperscript{1}The Hong Kong Polytechnic University\quad
    \small \textsuperscript{2}University of Science and Technology Beijing\quad \\
    \small \textsuperscript{3}NLPR, MAIS, Institute of Automation, Chinese Academy of Sciences\quad 
    \small \textsuperscript{4}University of Chinese Academy of Sciences\quad \\
    \small \textsuperscript{5}XYZ Embodied AI\quad 
    \small \textsuperscript{6}Shandong University\quad
     \small \textsuperscript{7}Linketic\quad
     \small \textsuperscript{8}D-Robotics
}

\begin{document}
\maketitle

\blfootnote{$^\dagger$ Project Leader.}
\blfootnote{\textsuperscript{\faEnvelope[regular]} Corresponding Author.}

\begin{abstract}

\noindent
The development of embodied agents for complex commercial environments is hindered by a critical gap in existing robotics datasets and benchmarks, which primarily focus on household or tabletop settings with short-horizon tasks. 
To address this limitation, we introduce MarketGen, 
a scalable simulation platform with automatic scene generation for complex supermarket environments.
MarketGen features a novel agent-based Procedural Content Generation (PCG) framework.
It uniquely supports multi-modal inputs (text and reference images) and integrates real-world design principles to automatically generate complete, structured, and realistic supermarkets.
We also provide an extensive and diverse 3D asset library with a total of \textbf{1100+} supermarket goods and parameterized facilities assets.
Building on this generative foundation, we propose a novel benchmark for assessing supermarket agents, featuring two daily tasks in a supermarket: (1) \textbf{Checkout Unloading:} long-horizon tabletop tasks for cashier agents, and (2) \textbf{In-Aisle Item Collection:} complex mobile manipulation tasks for salesperson agents. 
We validate our platform and benchmark through extensive experiments, including the deployment of a modular agent system and successful sim-to-real transfer. 
MarketGen provides a comprehensive framework to accelerate research in embodied AI for complex commercial applications.
Our project page is available at \url{https://xuhu0529.github.io/MarketGen}.
\end{abstract}    
\section{Introduction}
\label{sec:intro}

\begin{figure*}[t]
    \centering
    \includegraphics[width=0.99\linewidth]{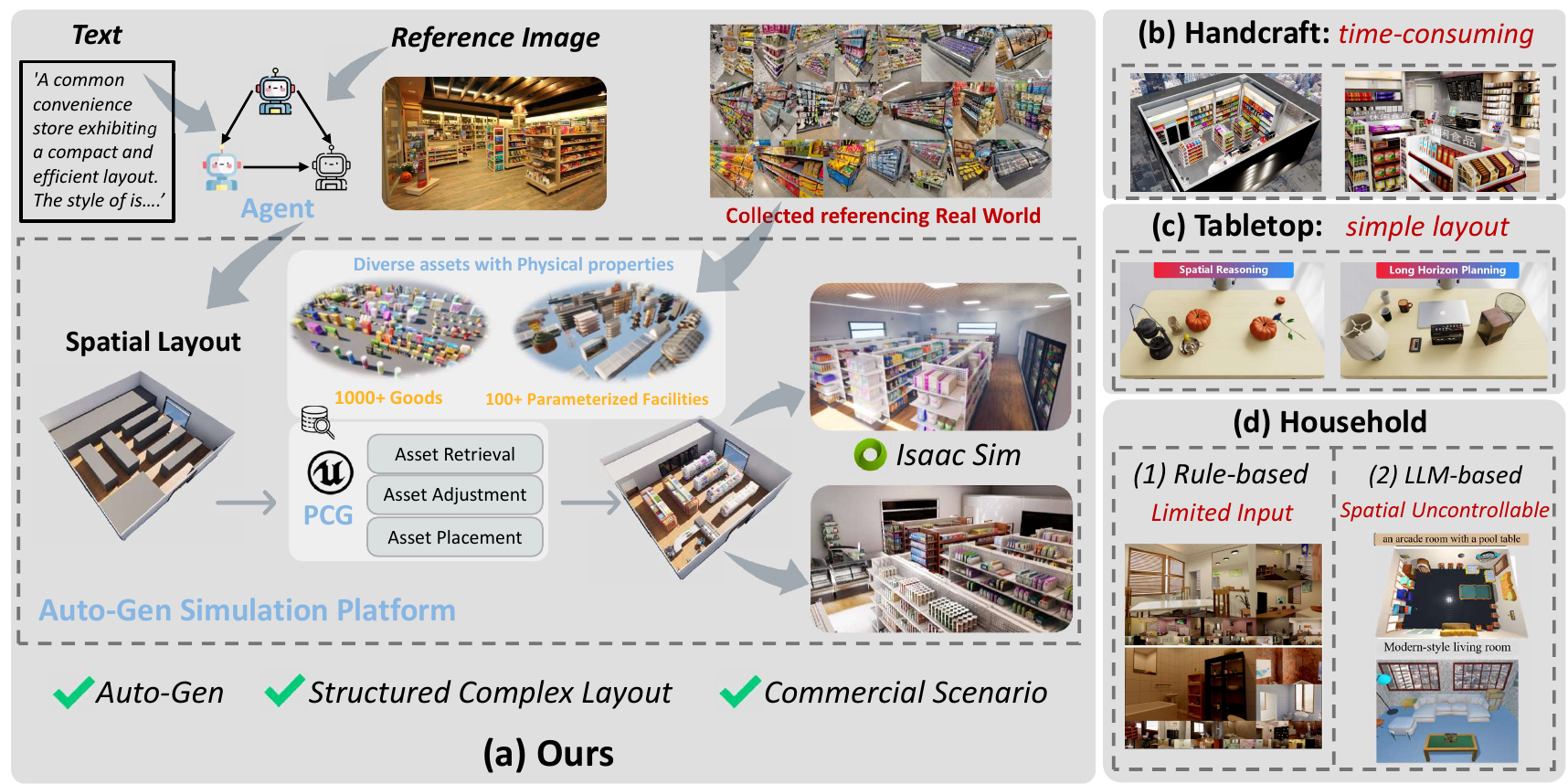}
    \caption{\textbf{Overview of MarketGen.} MarketGen features as a scalable simulation platform with auto-generated scenes for supermarket scenarios. It differs from previous platforms and methods: (b) Handcrafted supermarket scenes in GRUtopia~\cite{wang2024grutopia}, (c) Tabletop task generation~\cite{gao2025genmanip}, and (d) Rule-based~\cite {infinigen2024indoors,procthor,zhang2025agentworld} and LLM-based household scene generation~\cite{infiniteworld,Yang_2024_CVPR}.}
    \label{fig:teaser}
\end{figure*}

Recent advancements in embodied AI and robotic manipulation have highlighted the critical need for scalable, interactive simulation environments. 
These platforms are essential for developing, training, and collecting data for autonomous agents. 
While existing platforms have offered significant solutions, such as procedural scene generation\citep{zhang2025agentworld, infiniteworld, ssg, ai2thor, rcaregen} or the creation of task-specific manipulation datasets\citep{mimicgen, bigym, humanoidbench}, their focus has been limited to household and tabletop scenarios. 
A significant gap persists in addressing complex, large-scale commercial environments—such as supermarkets, hospitals, and factories. 
These scenarios represent key frontiers for the practical deployment of embodied technologies, yet they remain largely unexplored due to their unique scale and complexity.

A primary bottleneck in scaling to these commercial domains is the reliance on pre-designed, static scenes (Fig.~\ref{fig:teaser} (b)), which is time-consuming, labor-intensive, and fundamentally limits environmental diversity. 
Although automated scene generation solutions utilizing data-driven, LLM-based~\cite{ccelen2024design, sun2024layoutvlm, feng2024layoutgpt, Yang_2024_CVPR, fu2024anyhome,gao2025genmanip}, or PCG~\cite{infiniteworld, procthor, infinigen2024indoors} methods have emerged (Fig.~\ref{fig:teaser} (c) and (d)), they are also overwhelmingly designed for household or tabeltop scenarios. 
Consequently, these methods exhibit critical limitations when applied to commercial spaces, such as an inability to ensure controllable scene consistency, a lack of robust 3D spatial understanding (in LLM-only approaches), or highly restricted input modalities that fail to capture the complexity of a commercial layout.

To address these limitations, we introduce MarketGen, a novel, auto-gen simulation platform specifically designed for supermarket environments, as shown in Fig~\ref{fig:teaser}.
It is built upon NVIDIA’s Omniverse Isaac Sim~\cite{isaacsim} and Unreal Engine~\cite{UE}, thereby inheriting the advantages of both platforms, including the robust physics engine for robot simulation and photorealistic rendering capabilities.
To address the labor-intensive nature of handcrafting scenes, we propose the first automated scene generation system for supermarket scenarios. 
Our approach integrates an agent-based system with a PCG workflow, explicitly incorporating established principles of supermarket layout design. 
This hybrid methodology is designed to circumvent the limitations of standalone systems: it avoids the poor 3D spatial grounding inherent in LLM-only agents and the rigid input constraints of traditional PCG. 
The agent system is capable of interpreting diverse, high-level inputs, such as natural language descriptions or style-based reference images, to generate a functionally coherent scene layout.
The PCG workflow then automatically builds the full 3D scene based on the layout, populating it with the appropriate assets.

To overcome the challenge of limited scene resources in this domain, we also constructed an extensive and diverse 3D asset library. 
Collected referencing real-world supermarket items, this library includes over 1,000 high-fidelity models of common supermarket goods across 10 major categories (e.g., beverages, fruits and vegetables, dairy, snacks) and over 100 distinct supermarket facilities, including various shelving units, refrigerators, and checkout counters. 
Notably, we parametrically decomposed key facility assets, particularly shelving, allowing for flexible procedural control over properties like layer count, spacing, and modular combinations. 
This approach significantly expands the effective asset pool and enables the generation of highly varied, realistic interiors.

Finally, based on this platform, we propose a comprehensive benchmark for evaluating agent performance on tasks relevant to real-world supermarket operations. The benchmark task design includes: (1) \textbf{Checkout Unloading}, the long-horizon tabletop task, and (2) \textbf{In-aisle Item Collection}, the mobile manipulation task. 
These tasks are designed to meet real-world requirements for both static and dynamic agent capabilities. 
Concurrently, we designed a modular manipulation framework to support the zero-shot evaluation of visual prompting-based manipulation policies.

Our primary contributions are:
\begin{itemize}
    \item A scalable simulation platform with an automated scene generation system that combines a planning agent with a PCG workflow, enabling the controllable generation of diverse supermarket layouts from flexible user inputs.
    \item An extensive and diverse 3D asset library of over 1,000 high-fidelity supermarket products and 100+ parameterized facilities.
    \item A comprehensive benchmark for evaluating embodied agents on long-horizon and mobile manipulation tasks specific to supermarket operations, including a framework for zero-shot modular policy evaluation.
\end{itemize}



\begin{table*}[ht]
\centering
\resizebox{16cm}{!}{

\begin{tabular}{c|ccc|cccc|cc}
\toprule
\multirow{3}{*}{Name} & \multicolumn{3}{c}{\textbf{Assets}} & \multicolumn{4}{c}{\textbf{Scene Generation}} & \multicolumn{2}{c}{\textbf{Benchmark}} \\

\multirow{3}{*}{} & Num of    & \multirow{2}{*}{Parameterized}  & Physics & Commercial & \multirow{2}{*}{Auto-Gen} & \multirow{2}{*}{Text Input} & Reference     & Tabletop & Mobile \\
\multirow{3}{*}{} & Assets & \multirow{2}{*}{       } & Config  & Scenario & \multirow{2}{*}{        } & \multirow{2}{*}{      } & Image & Manipulation      & Manipulation \\
\midrule

Maniskill2~\cite{maniskill2}            & 2144     & \redcross        & \greencheck       & \redcross   & \redcross   & \redcross & \redcross & \greencheck & \redcross  \\
ProcTHOR~\cite{procthor}                & 3578     & \redcross        & \redcross     & \redcross & \greencheck   & \redcross & \redcross & \redcross & \redcross  \\
RLBench~\cite{rlbench}                  & 28       & \redcross      & \greencheck       & \redcross   & \redcross & \redcross & \redcross & \greencheck & \redcross    \\
BiGym~\cite{bigym}                      & $<$200   & \redcross      & \greencheck       & \redcross   & \redcross   & \redcross    & \redcross       & \greencheck & \redcross    \\
Behavior-1K~\cite{behavior1k}           & 5215     & \redcross        & \greencheck       & \redcross & \redcross   & \redcross & \redcross & \greencheck & \greencheck  \\
RoboCasa~\cite{robocasa}                & 2509     & \redcross      & \greencheck       & \redcross   & \redcross   & \redcross   & \redcross & \greencheck & \greencheck    \\
InfiniteWorld~\cite{infiniteworld}      & $>$10000 & \redcross      & \greencheck       & \redcross   & \greencheck   & \greencheck   & \redcross & \greencheck & \greencheck    \\
GRUtopia~\cite{wang2024grutopia}                & $\approx$25000 & \redcross      & \greencheck       & \greencheck   & \redcross   & \redcross   & \redcross           & \redcross & \greencheck    \\
AgentWorld~\cite{zhang2025agentworld}                              & $>$9000  & \redcross        & \greencheck       & \redcross    & \greencheck   & \redcross   & \redcross   & \greencheck   & \greencheck  \\
\midrule
MarketGen & $>$1100 & \greencheck     & \greencheck     & \greencheck     & \greencheck     & \greencheck     & \greencheck     & \greencheck     & \greencheck \\
\bottomrule
\end{tabular}
}
\caption{Comparison of robotic simulation platforms with other platforms in terms of asset properties, scene generation, and benchmark. MarketGen distinguishes itself by providing the first platform focusing on supermarket scenarios with a complete framework from automated scene generation to embodied task benchmarking and evaluation.}
\label{tab:sim_compare}
\end{table*}

\section{Related Works}
\label{sec:related_works}






\subsection{Simulation Platform for Embodied AI}
The rapid development of simulators is progressively transitioning from general-purpose functionality to high-fidelity realism. To reduce the sim-to-real gap, simulators need to ensure the realism of both the physics and the rendering and meet the diversity of the scenarios, assets, and tasks they provide. 
To achieve these goals, a variety of simulation platforms have emerged. 
ARNOLD~\cite{gong2023arnold}, VLMbench~\cite{zheng2022vlmbench}, Habitat~\cite{savva2019habitat}, ManiSkill3~\cite{tao2024maniskill3}, and ClevrSkills~\cite{haresh2024clevrskills} focus on language-guided task learning in realistic 3D environments, aiming to advance robotic manipulation and human-robot interaction research.
Furthermore, Behavior-1k~\cite{li2023behavior} and GRUtopia~\cite{wang2024grutopia} increase scene complexity by simulating human-like activities in pre-designed scenes, and Robacasa~\cite{robocasa} emphasizes household robotics with large-scale object interactions.
However, manually constructing high-fidelity and diverse scene environments for simulation is an exceedingly time-consuming and labor-intensive process, posing a significant bottleneck to large-scale experimentation and evaluation.
AgentWorld~\cite{zhang2025agentworld} integrate procedural scene generation to create large-scale interactive environments.
Although these existing works have designed and realized high-fidelity interactive simulation platforms, they focus on household scenarios and none of them explore to construct a simulation platform for the commercial environments, such as supermarket and grocery stores. 
MarketGen distinguishes itself by integrating procedural scene construction for supermarket generation with an agent system for layout generation, offering a reliable and scalable simulation platform for commercial robot landing.

\subsection{3D Indoor Scene Generation}
The task of 3D indoor scene generation is typically formulated as a layout prediction problem, where objects are represented by bounding boxes and semantic labels~\cite{fu20213d,paschalidou2021atiss,sun2024layoutvlm}. 
Data-driven generative methods~\cite{paschalidou2021atiss,tang2023diffuscene,yang2024physcene} often trained on large-scale datasets such as 3D-FRONT~\cite{fu20213d}, can learn to produce realistic, coarse-level scene layouts. However, these methods are often constrained by the limited variety and level-of-detail present in their training data.
LLM-based methods~\cite{ccelen2024design, sun2024layoutvlm, feng2024layoutgpt, Yang_2024_CVPR, fu2024anyhome} leverage the strengths of large language models to generate more detailed and contextually relevant indoor environments. 
Due to the poor spatial reasoning capability of LLMs, these methods suffer from hallucinations and show inconsistencies in object placements and geometric arrangements.
Prior works~\cite{infiniteworld, procthor, infinigen2024indoors} also explored procedural generation with primitive methods or large language models. 
However, most of these models focus on scene generation in household scenarios. To this end, we integrate an agent-based system with a PCG workflow, explicitly incorporating established principles of supermarket layout design.

\section{Simulation Platform}
\label{sec:methods}

We introduce the MarketGen simulation platform, which features two components: \textbf{3D Asset Library} and \textbf{Automatic Scene Generation}. Our system can automatically generate various structured supermarket scenes with diverse assets.
We compare the MarketGen simulation platform with popular platforms in Tab.~\ref{tab:sim_compare}.

\subsection{3D Asset Library}


To enable diverse scene construction, our simulation platform integrates a comprehensive collection of supermarket 3D assets, totally about 100+ basic facility assets (shelves, refrigerators, etc.) and 1000+ commodity assets. 
This wide array of assets covers a broad spectrum of goods typically found in a supermarket, varying in size, shape, and visual characteristics. 
See Fig.~\ref{fig:teaser} for an illustration of these assets.

\noindent \textbf{Rigid assets with annotations.}
The categories of the assets cover a wide range of supermarket needs, including fresh produce, beverages, packaged goods, .etc.
For each asset, we generate structured annotations via prompting Gemini-2.5-Pro, including object description, physical properties (scale, mass, friction), and semantic properties (category, color, material).

\noindent \textbf{Articulated Assets.}
For assets that necessitate dynamic interaction and state changes, such as the refrigerator doors or knobs on appliances (e.g., coffee machines), we
manually annotated the articulated constraints.

\noindent \textbf{Parameterized Facilities.}
Drawing from analyses of real-world supermarket facilities,
we deconstruct key infrastructure, particularly shelving systems, into their minimal constituent units. 
As shown in Fig.~\ref{fig:pcg}, these base components include elements like horizontal shelf boards, back panels, base supports, and side panels. 
By programmatically adjusting parameters, such as the number of vertical tiers, the spacing, and the unit's depth and length, the system can dynamically assemble a wide variety of shelving configurations (e.g., standard gondolas, wall units, end-caps). 
This parametric method not only ensures that the generated scenes possess a high degree of realism reflecting true-to-life layouts but also exponentially expands the combinatorial possibilities, significantly increasing the scale and diversity of the generated scenes.

\noindent \textbf{Visual Material Configuration.}
To enhance data augmentation for sim-to-real transfer, our platform also features a diverse library of high-fidelity Physically Based Rendering (PBR) materials. This library facilitates scene generalization by covering a wide range of common materials and textures in supermarket scenarios.
The collection is designed to represent diverse supermarket styles and aesthetics. For instance, for primary architectural surfaces such as walls and floors, we provide a selection of materials, including marble, brickwork, and various wood planks. Similarly, the visual characteristics of key fixtures, such as shelving units, can be programmatically adjusted by applying different materials, including various wood grains or metallic finishes.

\begin{figure*}[t]
    \centering
    \includegraphics[width=0.99\linewidth]{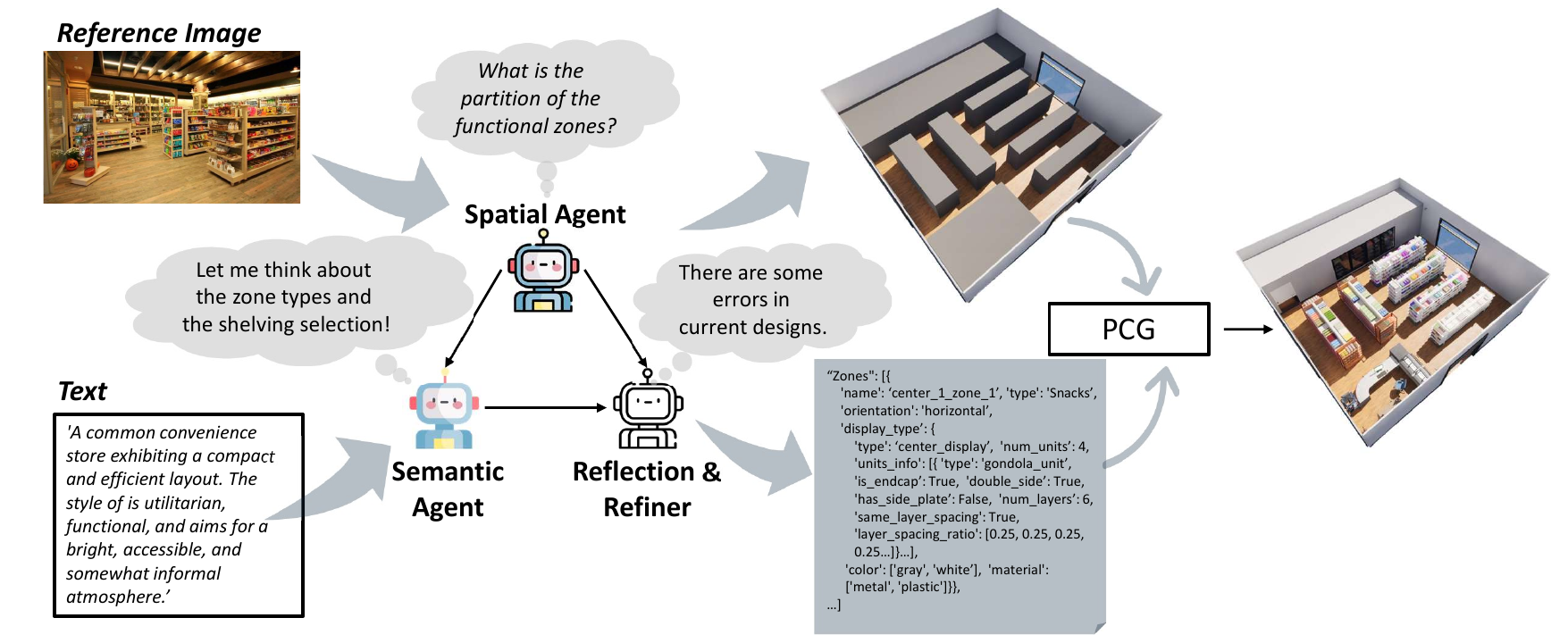}
    \caption{\textbf{The Pipeline of Automatic Scene Generation.} The agent system first generates a structured spatial layout and semantic info from the input text and reference image. Then the PCG workflow will finish scene construction.}
    \label{fig:agent}
\end{figure*}

\subsection{Automatic Scene Generation}
To address the challenge of time-consuming manual scene design, we first develop an agent-based system for the automated generation of supermarket layouts. 
This system serves as the high-level \textit{brain} that designs a coherent and realistic supermarket layout, which is then passed to the PCG workflow for 3D scene construction. The pipeline is illustrated in Fig.~\ref{fig:agent}.

\begin{figure}[t]
    \centering
    \includegraphics[width=0.99\linewidth]{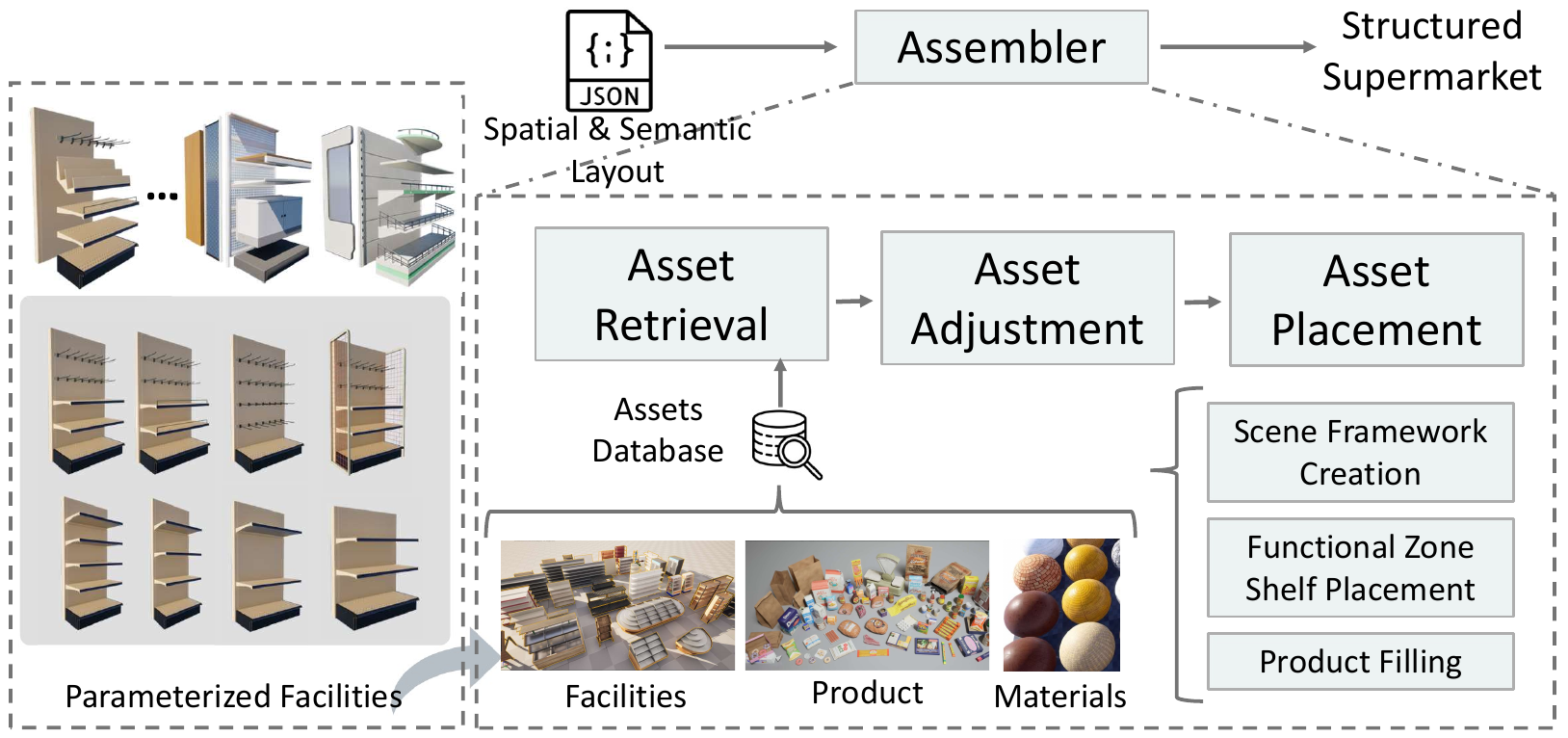}
    \caption{PCG Workflow with parameterized facilities.}
    \label{fig:pcg}
    \vspace{-0.5cm}
\end{figure}

\noindent \textbf{Agent System for Layout Generation.}
The system consists of three components: a Spatial Agent for functional zone planning, a Semantic Agent for asset specification, and a Reflection\&Refiner module for iterative plan correction.
\begin{itemize}
    \item \textbf{Spatial Agent} is responsible for parsing the overarching spatial division of the supermarket into distinct functional areas from text or reference image inputs. 
    Since LLMs inherently lack the capability to directly regress plausible spatial coordinates, we leverage the structured nature of supermarket layouts. 
    The LLM is tasked with predicting the parameters for pre-defined region division algorithms, i.e., Binary Space Partitioning (BSP), which then execute the spatial partitioning, ensuring the resulting layout conforms to realistic architectural constraints.
    \item \textbf{Semantic Agent} analyzes the specific requirements for shelving and facilities within each functional area.
    It determines the necessary asset properties, such as the style (e.g., with or without endcaps, single vs. double-sided), parametric attributes (e.g., number of layers, layer spacing), and visual characteristics (e.g., color, material).
    \item \textbf{Reflection \& Refiner} allows the LLM to reflect on the spatial arrangement, identify and correct errors in functional adjacencies or types, and refine the plan before it is finalized. 
    Since LLMs/VLMs always struggle with spatial reasoning, we employ a visual prompting strategy~\cite{sun2024layoutvlm}, where the rendered 2D layout with semantic and spatial labels from the initial plan is feed into the LLMs.
\end{itemize}


\noindent \textbf{PCG Workflow.}
With the layout serving as the blueprint for the overall scene structure, the PCG system automatically instantiates and configures these components. As shown in Fig.~\ref{fig:pcg}, this process consists of three primary stages: Asset Retrieval, Adjustment, and Placement.
\begin{itemize}
    \item \textbf{Asset Retrieval} initiates by parsing the semantic information specified in the 2D layout. This semantic data, which defines functional zones (e.g., \textit{produce}, \textit{dairy}, \textit{checkout}), is used to perform a matching retrieval query against our comprehensive Assets Database.
    \item \textbf{Asset Adjustment} adjusts the parameterized models of shelves and other supermarket facilities based on parameters defined in the layout, such as spatial dimensions, shelf placement types (e.g., \textit{wall-unit, gondola, with-endcap}), and the number of vertical layers.
    \item \textbf{Asset Placement} features more than Framework Creation and Zone Shelf Placement. Critically, it also includes an automated Product Filling procedure to contextually populate the shelves with appropriate products according to the functional zone's semantic type.
\end{itemize}

\noindent
This workflow can significantly streamlines the scene setup process from the layout, ensuring a visually rich and functionally complete simulated supermarket environment.




\begin{figure*}[t]
    \centering
    \includegraphics[width=0.99\linewidth]{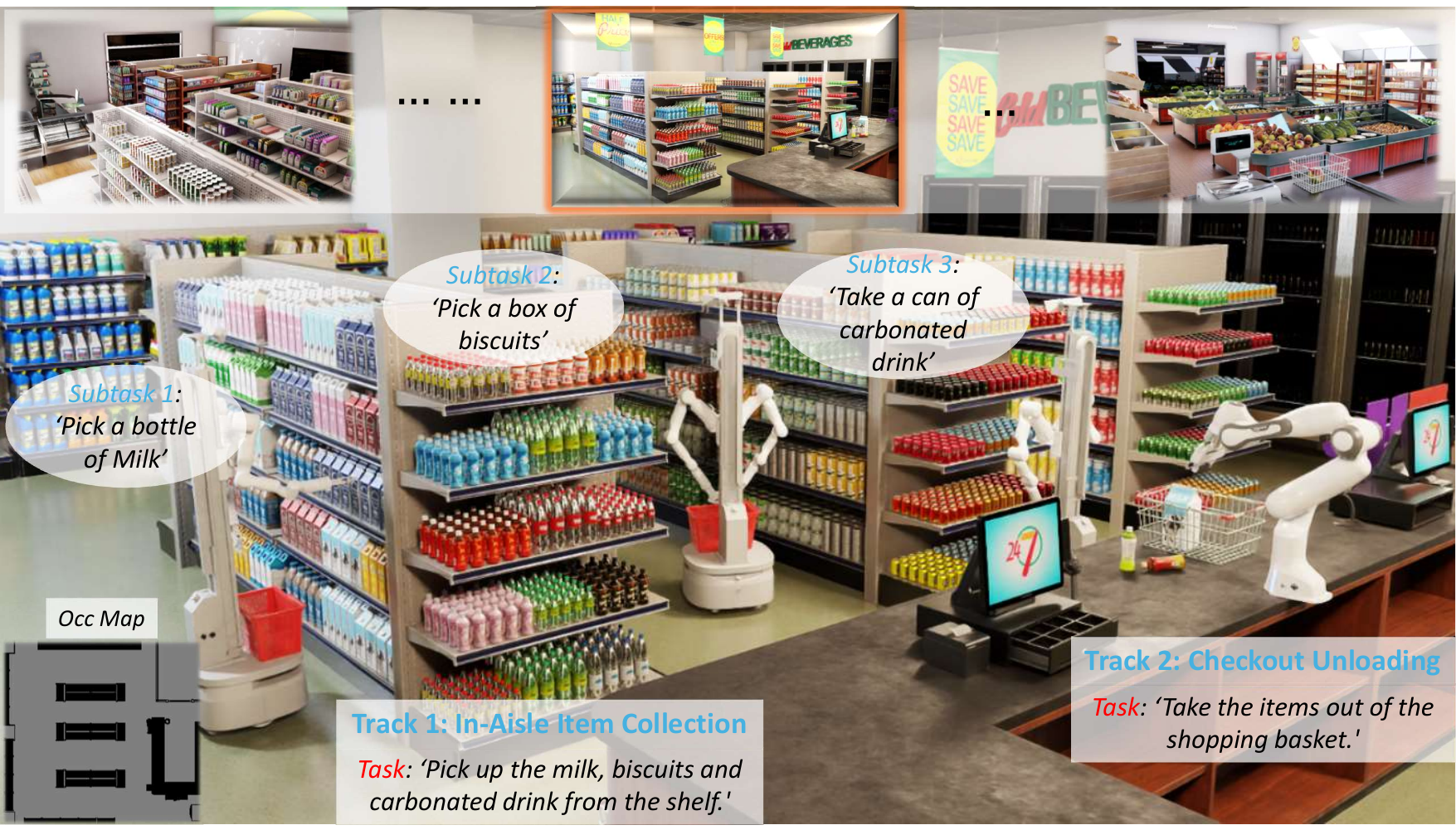}
    \caption{An overview of our benchmark. There are two tracks: \textbf{Checkout Unloading} for tabletop manipulation tasks and \textbf{In-Aisle Item Collection} for mobile manipulation tasks.}
    \label{fig:benchmark}
\end{figure*}

\section{Benchmark for Supermarket Agents}
\label{sec:bench}

To evaluate the capabilities of embodied robots within supermarket environments, we establish a dedicated benchmark primarily focused on long-horizon manipulation tasks.
The design of this benchmark is grounded in practical applications, drawing inspiration directly from the daily operational duties performed by human supermarket staff.

\subsection{Scene \& Task Setting}
To benchmark agent performance, we establish two distinct benchmark tracks, each derived from common yet fundamentally different daily supermarket operations:
\begin{itemize}
    \item \textbf{Checkout Unloading} is modeled on the duties of a cashier. It is primarily a stationary tabletop manipulation task. The agent is positioned at a checkout counter and is required to retrieve (pick up) a variety of items from a shopping basket, simulating the unloading step necessary for subsequent actions like scanning.
    \item \textbf{In-Aisle Item Collection} is inspired by the role of a clerk or staff member fulfilling an order. This task constitutes a mobile manipulation challenge. It requires the agent to navigate through the shopping area to various locations, successfully grasp different target products from the shelves, and collect them by placing them into a shopping basket. This scenario inherently tests a broader set of skills, integrating navigation, long-range perception, and multi-object grasping.
\end{itemize}

We generate 10 unique benchmark scenes and randomly sample 100 episodes for each benchmark for evaluation. Each episode has 2 to 4 target objects to be picked up.



\noindent{\textbf{Task Sampling.}}
To ensure the feasibility of all sampled tasks, we must validate that a valid solution exists. This involves sampling an agent's start position and several target objects, then confirming that the task is solvable. A task is deemed solvable only if two conditions are met: 1) a collision-free navigation path exists from the start position to the vicinity of the objects, and 2) the target objects are kinematically reachable and graspable from that location.

To efficiently verify the navigation constraint, we first pre-process each scene by generating a 2D occupancy map.
We then utilize this occupancy map to search for a collision-free path (e.g., using an A* search) from a randomly sampled point within the traversable area to the target object's location, thereby guaranteeing navigational accessibility.
To satisfy the graspability constraint, we apply a selection heuristic. Only items located on the outermost, front-facing layer of a shelf are considered valid targets. This simplifying assumption ensures that the selected object is not occluded by other items, making it kinematically reachable for a grasping attempt.






\noindent\textbf{Evaluation metrics.} We evaluate the results using three widely adopted metrics: success rate (SR), path length (PL) and success rate weighted by path length (SPL). 

\begin{itemize}
    \item \textbf{SR:} Success is defined as the agent correctly meeting the goal conditions. For $M$ goal conditions, achieving one condition yields a score of $1/M$.
    \begin{equation}
        \text{SR Score} = \frac{1}{N}\sum_{i=1}^N \frac{1}{M}\sum_{j=1}^M1{\{\text{goal condition}\}}\
    \end{equation}
    
    \item \textbf{SPL:} SPL score is calculated as follows:
    \begin{equation}
        \text{SPL Score} = \frac{1}{N}\sum_{i=1}^{N}S_i\frac{l_i}{\max(p_i,l_i)}
    \end{equation}
    where $N$ is the number of test episodes, $l_i$ is the shortest path distance from the start position to the target position, and $p_i$ is the length of the path actually taken. $S_i$ denotes the success rate in the episode. 
\end{itemize}

\noindent\textbf{Robot Setups.}
For the \textbf{Checkout Unloading} task, we use the Franka robot equipped with an RGB-D camera for perception. For the \textbf{In-Aisle Item Collection} task, we use the dual-arm Realman robot with a mobile and lifting base.

\subsection{Modular Manipulation System}



Recent modular methods~\cite{moka, copa, google2024pivot, robopoint, gao2025genmanip} leverage mark-based visual prompting in LLMs or foundation models to achieve strong generalization in real-world tasks.
Inspired by GENMANIP~\cite{gao2025genmanip}, we additionally design a modular policy for complex mobile manipulation tasks. Our strategy includes three core components: an Affordance Generator, an Action Executor, and a Long Horizon Planner.

\noindent \textbf{Affordance Generator.}
We begin by employing the Segment Anything Model (SAM)~\cite{kirillov2023sam} to generate masks for the observed scene, with each mask assigned a unique number identifier. Then we utilize Set-of-Mark (SoM)~\cite{yang2023som} to prompt VLM to select the target object’s mask based on instructions. 
We employ AnyGrasp~\cite{fang2023anygrasp} to generate candidate grasp poses for the target object from SoM mask, and then perform collision checking and heuristics to filter out the target grasp pose.

\noindent \textbf{Action Executor.}
After generating the grasp pose, we employ cuRobo~\cite{sundaralingam2023curobo}, a high-performance motion planning library, to compute a collision-free trajectory for the manipulator.
For \textbf{In-Aisle Item Collection} task, to streamline the agent's navigation and exploration process for target commodities, we provide the system with a prior environmental knowledge. 
This includes a 2D functional layout map of the supermarket and the 2D positional data corresponding to the target item's category (not the instance).
We contend that this is a reasonable simplification, as it mirrors real-world operational scenarios where staff routinely possess this knowledge of the supermarket's categorical organization.
Providing this information enables the agent to avoid exhaustive, blind search strategies and proceed directly to the relevant area.

\noindent \textbf{Long Horizon Planner.}
The Long-Horizon Planner module employs LLMs or VLMs to manage high-level task decomposition and completion checking.
The task decomposition function analyzes the instruction, historical context, and the current environmental state to break down the long-horizon objective into a logical sequence of discrete sub-tasks.
Furthermore, the planner incorporates a completion feedback loop. This will evaluate the execution status of the current sub-task, determines its success or failure, and assesses whether recovery procedures are necessary to remediate any faults from preceding actions, thus ensuring robust task progression.
\section{Experiments}
\label{sec:exps}

\begin{figure*}[t]
    \centering
    \includegraphics[width=0.99\linewidth]{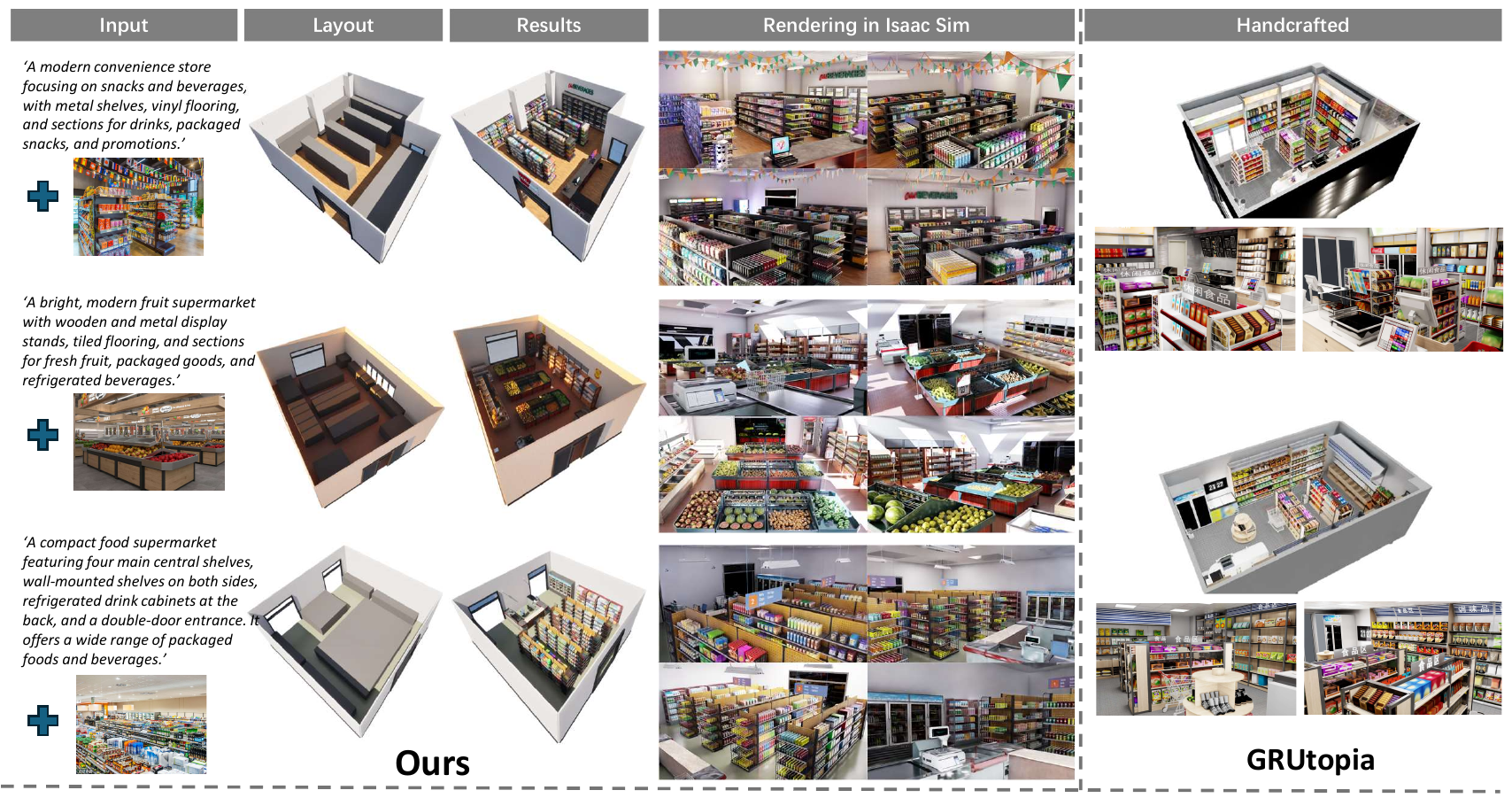}
    \caption{Results from our automatic scene generation pipeline. We can achieve a level of fidelity and logical coherence comparable to handcrafted scenes in GRUtopia. }
    \label{fig:scene_gen_res}
\end{figure*}

\subsection{Scene Generation Results}

We conduct a qualitative evaluation to demonstrate the efficacy, realism, and diversity of our automated scene generation pipeline.
As shown in Fig.~\ref{fig:scene_gen_res}, our system successfully generates a wide variety of complex and densely populated supermarket scenes. 
The results achieve a level of fidelity and logical coherence comparable to meticulously handcrafted environments, such as those in GRUtopia~\cite{wang2024grutopia}.

\noindent
The spatial plausibility is a result of our hybrid pipeline, which explicitly embeds supermarket design principles within both the agent system and PCG workflow.
This ensures that functional zones from produce sections and dairy cases to checkout counters and main aisles are arranged in a structured and realistic manner.
Furthermore, the parametric shelving allows for significant structural variation in aisle configuration, while the rich repository of over 1,000 product assets ensures that these aisles are populated with high visual diversity.
Finally, as demonstrated in the rendered views of the simulator, the high-fidelity visual quality is crucial for minimizing the visual sim-to-real gap.

\begin{table}[t]
\centering
\caption{Quantitative results of two benchmark tracks.}
\label{tab: main results}
\resizebox{0.48\textwidth}{!}{%
\begin{tabular}{cccccc}
\toprule
 \multirow{2}{*}{MLLM} & \multicolumn{1}{c}{Checkout Unloading} & \multicolumn{3}{c}{In-Aisle Item Collection} & Overall \\ 
\cmidrule(lr){2-2} \cmidrule(lr){3-5} \cmidrule(lr){6-6} 
 & \multicolumn{1}{c}{SR(\%)} & SR(\%) & SPL(\%) & PL($m$) & SR(\%) \\ \midrule
 GPT-4o & 14.22  & 8.33 & 4.09 & \textbf{14.95} & 11.28\\
 Claude-Sonnet-4.5 & \textbf{19.77} & 6.66 & 3.84 & 11.43 & 13.22 \\
 Gemini-2.5-Pro & 17.32 & \textbf{10.41} & \textbf{5.29} & 12.20 & \textbf{13.87} \\
 Qwen3-VL-Plus & 15.20 & 9.20 & 5.08 & 12.44 & 12.20 \\
\bottomrule
\end{tabular}
}
\vspace{-0.5cm}
\end{table} 

\subsection{Modular Manipulation Results}

\begin{figure*}[t]
    \centering
    \includegraphics[width=0.99\linewidth]{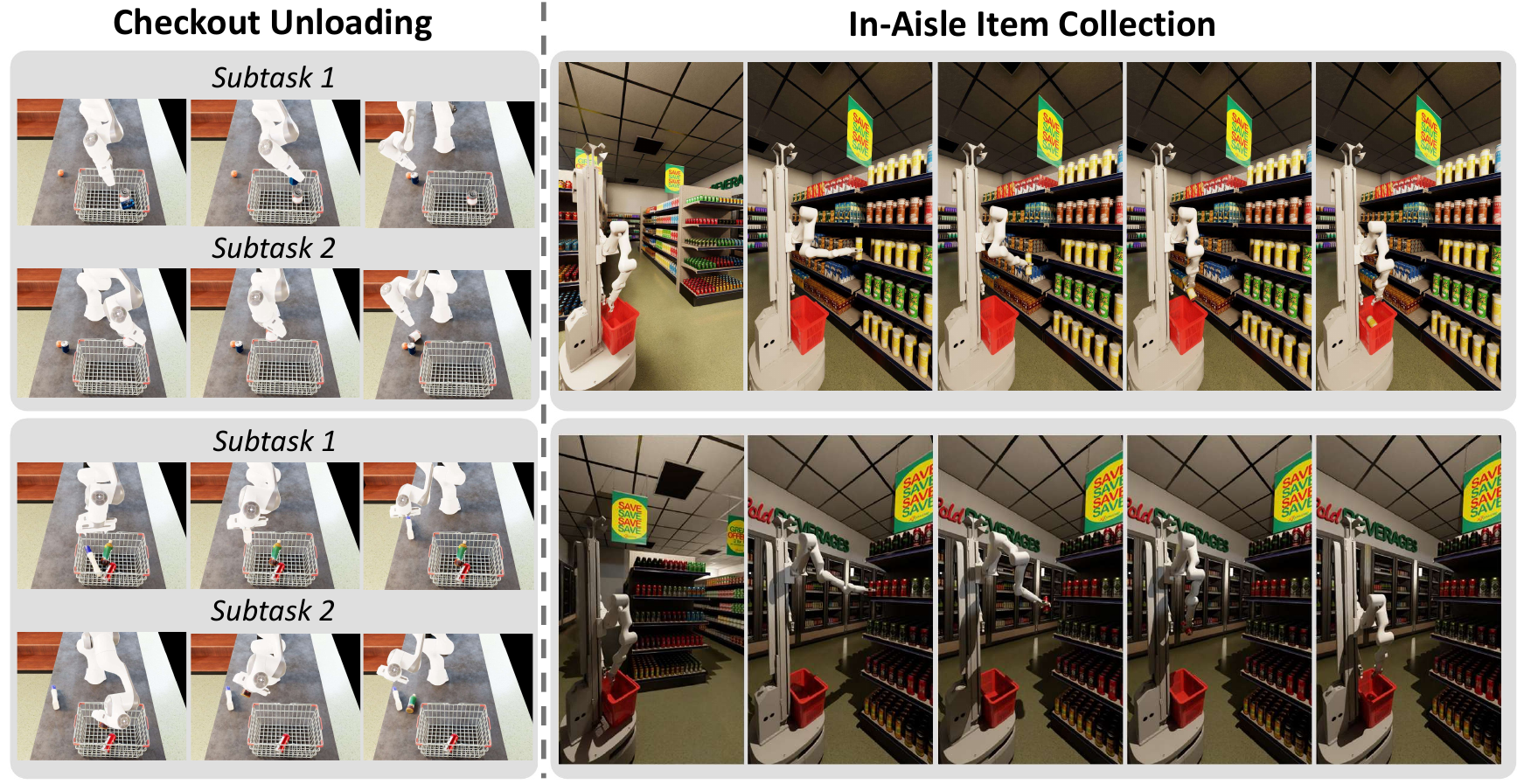}
    \caption{Qualitative results for two benchmark tracks.}
    \label{fig:benchmark_exps}
    \vspace{-0.5cm}
\end{figure*}

\begin{figure}[t]
    \centering
    \includegraphics[width=0.99\linewidth]{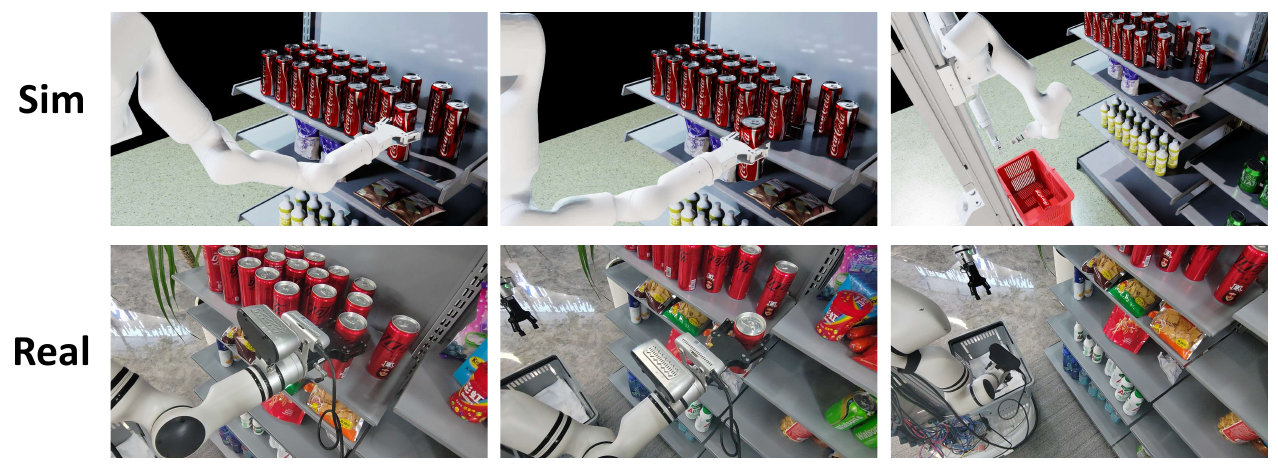}
    \caption{A sim-to-real transfer example.}
    \label{fig:sim_real}
\end{figure}



We evaluate the performance of several representative Multi-modal Large Language Models (MLLMs) on our two benchmark tracks: Checkout Unloading and In-Aisle Item Collection.
As shown in Tab.~\ref{tab: main results}, the results highlight the significant challenge posed by long-horizon, high-clutter tasks in realistic supermarket scenarios. 
While Gemini-2.5-Pro achieves the highest overall Success Rate (SR), its performance remains below 15\%, indicating that the modular methods struggle with the complexity of these tasks.

Our analysis identifies several key difficulties. 
As shown in Fig.~\ref{fig:benchmark_exps}, in the Checkout Unloading benchmark, items are freely placed in the shopping basket, resulting in occlusion and complex inter-object relationships. This requires the agent to perform sophisticated reasoning to perceive and successfully grasp target items. 
In the In-Aisle Item Collection benchmark, the agent must not only navigate to the correct location but also accurately locate a specific item among densely arranged, visually similar products on a shelf. 
The narrow aisles and packed shelving further increase the task difficulty, introducing a high risk of collision with the shelves or adjacent items during manipulation.

These low success rates suggest that relying solely on MLLMs for high-level planning and separate affordance prediction models for manipulation is insufficient for these complex environments. 
We observe that these models' dependence on intermediate representations, rather than end-to-end optimization, often poses challenges for generating optimal paths. 
This can lead to suboptimal or inefficient task execution, underscoring the need for future advancements in integrating end-to-end learning approaches to enhance the overall performance of robotic manipulation systems in such complex, real-world scenarios.









\subsection{Real-world Experiment}

To verify our simulation platform's capability for sim-to-real transfer, we conduct a direct comparative experiment focusing on the performance of the affordance prediction model. 
For the evaluation, we deploy this model in the real world and select four common objects representative of distinct shapes (e.g., box, can, bottle, bag) and arrange them in a real-world layout, as shown in Fig.~\ref{fig:sim_real}. 
A similar scene, replicating the real environment's target object and shelf types, positions, and relative layout, is then constructed in our platform using the parameterized assets. 

We execute 20 grasp experiments for each of the four object shapes under varied settings (including different robot's positions and object categories) in both the real and simulation environments.
Summarized in Tab.~\ref{tab: real exps}, the results show a strong correlation between the model's success rates in simulation and real world. 
This close performance alignment demonstrates that the realistic environmental layouts, accurate asset physics, and high-fidelity rendering quality of our simulation platform provide a solid foundation for bridging the sim-to-real gap.

\begin{table}[t]
\centering
\caption{Evaluation of the performance of the affordance prediction model on 4 different object shapes. The metric is selected as Success Rate (SR).}
\label{tab: real exps}
\resizebox{0.35\textwidth}{!}{%
\begin{tabular}{ccccc}
\toprule
  & Bottles & Cans & Boxes & Bags \\ \midrule
 Sim & 55.0\% & 45.0\% & 25.0\% & 0.0   \\
 Real & 50.0\% & 45.0\% & 15.0\% & 0.0  \\
\bottomrule
\end{tabular}
}
\vspace{-0.5cm}
\end{table}

\section{Conclusion}
\label{sec:conclusion}

We present MarketGen, a scalable simulation platform with automatic scene generation for complex supermarket environments. 
At its core is a novel agent-based Procedural Content Generation (PCG) framework, which supports multi-modal inputs to automatically generate complete, structured, and realistic supermarket scenes.
We also present a benchmark for assessing the daily tasks of supermarket agents. Experiments show that modular methods struggle with the complexity of these tasks in supermarket scenarios, underscoring the need for future advancements in integrating end-to-end learning methods to enhance the overall performance.
Finally, the close performance alignment in the sim-to-real transfer experiments demonstrates that the MarketGen simulation platform provides a solid foundation for bridging the sim-to-real gap.
{
    \small
    \bibliographystyle{ieeenat_fullname}
    \bibliography{main}

\begin{thebibliography}{41}
\providecommand{\natexlab}[1]{#1}
\providecommand{\url}[1]{\texttt{#1}}
\expandafter\ifx\csname urlstyle\endcsname\relax
  \providecommand{\doi}[1]{doi: #1}\else
  \providecommand{\doi}{doi: \begingroup \urlstyle{rm}\Url}\fi

\bibitem[{\c{C}}elen et~al.(2024){\c{C}}elen, Han, Schindler, Van~Gool, Armeni, Obukhov, and Wang]{ccelen2024design}
Ata {\c{C}}elen, Guo Han, Konrad Schindler, Luc Van~Gool, Iro Armeni, Anton Obukhov, and Xi Wang.
\newblock I-design: Personalized llm interior designer.
\newblock \emph{arXiv preprint arXiv:2404.02838}, 2024.

\bibitem[Chen et~al.(2025)Chen, Ye, Dingwani, Fazli, Seifi, and Bhattacharjee]{rcaregen}
Shuaixing Chen, Ruolin Ye, Saurabh Dingwani, Pooyan Fazli, Hasti Seifi, and Tapomayukh Bhattacharjee.
\newblock Rcaregen: An interface for scene and task generation in rcareworld.
\newblock In \emph{Proceedings of the 2025 ACM/IEEE International Conference on Human-Robot Interaction}, pages 1274--1278, 2025.

\bibitem[Chernyadev et~al.()Chernyadev, Backshall, Ma, Lu, Seo, and James]{bigym}
Nikita Chernyadev, Nicholas Backshall, Xiao Ma, Yunfan Lu, Younggyo Seo, and Stephen James.
\newblock Bigym: A demo-driven mobile bi-manual manipulation benchmark.
\newblock In \emph{8th Annual Conference on Robot Learning}.

\bibitem[Deitke et~al.(2022)Deitke, VanderBilt, Herrasti, Weihs, Salvador, Ehsani, Han, Kolve, Farhadi, Kembhavi, and Mottaghi]{procthor}
Matt Deitke, Eli VanderBilt, Alvaro Herrasti, Luca Weihs, Jordi Salvador, Kiana Ehsani, Winson Han, Eric Kolve, Ali Farhadi, Aniruddha Kembhavi, and Roozbeh Mottaghi.
\newblock {ProcTHOR: Large-Scale Embodied AI Using Procedural Generation}.
\newblock In \emph{NeurIPS}, 2022.
\newblock Outstanding Paper Award.

\bibitem[Engine(2018)]{UE}
Unreal Engine.
\newblock Unreal engine.
\newblock \emph{Retrieved from Unreal Engine: https://www. unrealengine. com/en-US/what-is-unreal-engine-4}, 2018.

\bibitem[Fang et~al.(2023)Fang, Wang, Fang, Gou, Liu, Yan, Liu, Xie, and Lu]{fang2023anygrasp}
Hao-Shu Fang, Chenxi Wang, Hongjie Fang, Minghao Gou, Jirong Liu, Hengxu Yan, Wenhai Liu, Yichen Xie, and Cewu Lu.
\newblock Anygrasp: Robust and efficient grasp perception in spatial and temporal domains.
\newblock \emph{IEEE Transactions on Robotics}, 39\penalty0 (5):\penalty0 3929--3945, 2023.

\bibitem[Feng et~al.(2024)Feng, Zhu, Fu, Jampani, Akula, He, Basu, Wang, and Wang]{feng2024layoutgpt}
Weixi Feng, Wanrong Zhu, Tsu-jui Fu, Varun Jampani, Arjun Akula, Xuehai He, Sugato Basu, Xin~Eric Wang, and William~Yang Wang.
\newblock Layoutgpt: Compositional visual planning and generation with large language models.
\newblock \emph{NeurIPS}, 2024.

\bibitem[Fu et~al.(2021)Fu, Cai, Gao, Zhang, Wang, Li, Zeng, Sun, Jia, Zhao, et~al.]{fu20213d}
Huan Fu, Bowen Cai, Lin Gao, Ling-Xiao Zhang, Jiaming Wang, Cao Li, Qixun Zeng, Chengyue Sun, Rongfei Jia, Binqiang Zhao, et~al.
\newblock 3d-front: 3d furnished rooms with layouts and semantics.
\newblock In \emph{ICCV}, 2021.

\bibitem[Fu et~al.(2024)Fu, Wen, Liu, and Sridhar]{fu2024anyhome}
Rao Fu, Zehao Wen, Zichen Liu, and Srinath Sridhar.
\newblock Anyhome: Open-vocabulary generation of structured and textured 3d homes.
\newblock In \emph{ECCV}, 2024.

\bibitem[Gao et~al.(2025)Gao, Chen, Yang, Chen, Tian, Li, Huang, Wang, Wang, and Pang]{gao2025genmanip}
Ning Gao, Yilun Chen, Shuai Yang, Xinyi Chen, Yang Tian, Hao Li, Haifeng Huang, Hanqing Wang, Tai Wang, and Jiangmiao Pang.
\newblock Genmanip: Llm-driven simulation for generalizable instruction-following manipulation.
\newblock In \emph{Proceedings of the Computer Vision and Pattern Recognition Conference}, pages 12187--12198, 2025.

\bibitem[Gong et~al.(2023)Gong, Huang, Zhao, Geng, Gao, Wu, Ai, Zhou, Terzopoulos, Zhu, et~al.]{gong2023arnold}
Ran Gong, Jiangyong Huang, Yizhou Zhao, Haoran Geng, Xiaofeng Gao, Qingyang Wu, Wensi Ai, Ziheng Zhou, Demetri Terzopoulos, Song-Chun Zhu, et~al.
\newblock Arnold: A benchmark for language-grounded task learning with continuous states in realistic 3d scenes.
\newblock In \emph{Proceedings of the IEEE/CVF International Conference on Computer Vision}, pages 20483--20495, 2023.

\bibitem[Gu et~al.(2023)Gu, Xiang, Li, Ling, Liu, Mu, Tang, Tao, Wei, Yao, et~al.]{maniskill2}
Jiayuan Gu, Fanbo Xiang, Xuanlin Li, Zhan Ling, Xiqiang Liu, Tongzhou Mu, Yihe Tang, Stone Tao, Xinyue Wei, Yunchao Yao, et~al.
\newblock Maniskill2: A unified benchmark for generalizable manipulation skills.
\newblock \emph{arXiv preprint arXiv:2302.04659}, 2023.

\bibitem[Haresh et~al.(2024)Haresh, Dijkman, Bhattacharyya, and Memisevic]{haresh2024clevrskills}
Sanjay Haresh, Daniel Dijkman, Apratim Bhattacharyya, and Roland Memisevic.
\newblock Clevrskills: Compositional language and visual reasoning in robotics.
\newblock In \emph{The Thirty-eight Conference on Neural Information Processing Systems Datasets and Benchmarks Track}, 2024.

\bibitem[Huang et~al.(2024)Huang, Lin, Hu, Wang, and Gao]{copa}
Haoxu Huang, Fanqi Lin, Yingdong Hu, Shengjie Wang, and Yang Gao.
\newblock Copa: General robotic manipulation through spatial constraints of parts with foundation models.
\newblock \emph{arXiv preprint arXiv:2403.08248}, 2024.

\bibitem[James et~al.(2020)James, Ma, Arrojo, and Davison]{rlbench}
Stephen James, Zicong Ma, David~Rovick Arrojo, and Andrew~J Davison.
\newblock Rlbench: The robot learning benchmark \& learning environment.
\newblock \emph{IEEE Robotics and Automation Letters}, 5\penalty0 (2):\penalty0 3019--3026, 2020.

\bibitem[Kirillov et~al.(2023)Kirillov, Mintun, Ravi, Mao, Rolland, Gustafson, Xiao, Whitehead, Berg, Lo, et~al.]{kirillov2023sam}
Alexander Kirillov, Eric Mintun, Nikhila Ravi, Hanzi Mao, Chloe Rolland, Laura Gustafson, Tete Xiao, Spencer Whitehead, Alexander~C Berg, Wan-Yen Lo, et~al.
\newblock Segment anything.
\newblock In \emph{Proceedings of the IEEE/CVF international conference on computer vision}, pages 4015--4026, 2023.

\bibitem[Kolve et~al.(2017)Kolve, Mottaghi, Han, VanderBilt, Weihs, Herrasti, Deitke, Ehsani, Gordon, Zhu, et~al.]{ai2thor}
Eric Kolve, Roozbeh Mottaghi, Winson Han, Eli VanderBilt, Luca Weihs, Alvaro Herrasti, Matt Deitke, Kiana Ehsani, Daniel Gordon, Yuke Zhu, et~al.
\newblock Ai2-thor: An interactive 3d environment for visual ai.
\newblock \emph{arXiv preprint arXiv:1712.05474}, 2017.

\bibitem[Li et~al.(2023{\natexlab{a}})Li, Zhang, Wong, Gokmen, Srivastava, Mart{\'\i}n-Mart{\'\i}n, Wang, Levine, Lingelbach, Sun, et~al.]{behavior1k}
Chengshu Li, Ruohan Zhang, Josiah Wong, Cem Gokmen, Sanjana Srivastava, Roberto Mart{\'\i}n-Mart{\'\i}n, Chen Wang, Gabrael Levine, Michael Lingelbach, Jiankai Sun, et~al.
\newblock Behavior-1k: A benchmark for embodied ai with 1,000 everyday activities and realistic simulation.
\newblock In \emph{Conference on Robot Learning}, pages 80--93. PMLR, 2023{\natexlab{a}}.

\bibitem[Li et~al.(2023{\natexlab{b}})Li, Zhang, Wong, Gokmen, Srivastava, Mart{\'\i}n-Mart{\'\i}n, Wang, Levine, Lingelbach, Sun, et~al.]{li2023behavior}
Chengshu Li, Ruohan Zhang, Josiah Wong, Cem Gokmen, Sanjana Srivastava, Roberto Mart{\'\i}n-Mart{\'\i}n, Chen Wang, Gabrael Levine, Michael Lingelbach, Jiankai Sun, et~al.
\newblock Behavior-1k: A benchmark for embodied ai with 1,000 everyday activities and realistic simulation.
\newblock In \emph{Conference on Robot Learning}, pages 80--93. PMLR, 2023{\natexlab{b}}.

\bibitem[Li et~al.(2022)Li, Guo, Liu, and Sun]{ssg}
Xinghang Li, Di Guo, Huaping Liu, and Fuchun Sun.
\newblock Embodied semantic scene graph generation.
\newblock In \emph{Conference on robot learning}, pages 1585--1594. PMLR, 2022.

\bibitem[Liu et~al.(2024)Liu, Fang, Abbeel, and Levine]{moka}
Fangchen Liu, Kuan Fang, Pieter Abbeel, and Sergey Levine.
\newblock Moka: Open-vocabulary robotic manipulation through mark-based visual prompting.
\newblock In \emph{First Workshop on Vision-Language Models for Navigation and Manipulation at ICRA 2024}, 2024.

\bibitem[Mandlekar et~al.(2023)Mandlekar, Nasiriany, Wen, Akinola, Narang, Fan, Zhu, and Fox]{mimicgen}
Ajay Mandlekar, Soroush Nasiriany, Bowen Wen, Iretiayo Akinola, Yashraj Narang, Linxi Fan, Yuke Zhu, and Dieter Fox.
\newblock Mimicgen: A data generation system for scalable robot learning using human demonstrations.
\newblock \emph{arXiv preprint arXiv:2310.17596}, 2023.

\bibitem[Nasiriany et~al.(2024{\natexlab{a}})Nasiriany, Maddukuri, Zhang, Parikh, Lo, Joshi, Mandlekar, and Zhu]{robocasa}
Soroush Nasiriany, Abhiram Maddukuri, Lance Zhang, Adeet Parikh, Aaron Lo, Abhishek Joshi, Ajay Mandlekar, and Yuke Zhu.
\newblock Robocasa: Large-scale simulation of everyday tasks for generalist robots.
\newblock \emph{arXiv preprint arXiv:2406.02523}, 2024{\natexlab{a}}.

\bibitem[Nasiriany et~al.(2024{\natexlab{b}})Nasiriany, Xia, Yu, Xiao, Liang, Dasgupta, Xie, Driess, Wahid, Xu, Vuong, Zhang, Lee, Lee, Xu, Kirmani, Zhu, Zeng, Hausman, Heess, Finn, Levine, and Ichter]{google2024pivot}
Soroush Nasiriany, Fei Xia, Wenhao Yu, Ted Xiao, Jacky Liang, Ishita Dasgupta, Annie Xie, Danny Driess, Ayzaan Wahid, Zhuo Xu, Quan Vuong, Tingnan Zhang, Tsang-Wei~Edward Lee, Kuang-Huei Lee, Peng Xu, Sean Kirmani, Yuke Zhu, Andy Zeng, Karol Hausman, Nicolas Heess, Chelsea Finn, Sergey Levine, and Brian Ichter.
\newblock Pivot: Iterative visual prompting elicits actionable knowledge for vlms.
\newblock 2024{\natexlab{b}}.

\bibitem[{NVIDIA}(2025)]{isaacsim}
{NVIDIA}.
\newblock Isaac sim 4.5 - robotics simulation and synthetic data generation.
\newblock \url{https://developer.nvidia.com/isaacsim}, 2025.

\bibitem[Paschalidou et~al.(2021)Paschalidou, Kar, Shugrina, Kreis, Geiger, and Fidler]{paschalidou2021atiss}
Despoina Paschalidou, Amlan Kar, Maria Shugrina, Karsten Kreis, Andreas Geiger, and Sanja Fidler.
\newblock Atiss: Autoregressive transformers for indoor scene synthesis.
\newblock In \emph{NeurIPS}, 2021.

\bibitem[Raistrick et~al.(2024)Raistrick, Mei, Kayan, Yan, Zuo, Han, Wen, Parakh, Alexandropoulos, Lipson, Ma, and Deng]{infinigen2024indoors}
Alexander Raistrick, Lingjie Mei, Karhan Kayan, David Yan, Yiming Zuo, Beining Han, Hongyu Wen, Meenal Parakh, Stamatis Alexandropoulos, Lahav Lipson, Zeyu Ma, and Jia Deng.
\newblock Infinigen indoors: Photorealistic indoor scenes using procedural generation.
\newblock In \emph{CVPR}, 2024.

\bibitem[Ren et~al.(2024)Ren, Li, Luo, Song, Chen, Liufu, Yang, Zheng, Xu, Huang, et~al.]{infiniteworld}
Pengzhen Ren, Min Li, Zhen Luo, Xinshuai Song, Ziwei Chen, Weijia Liufu, Yixuan Yang, Hao Zheng, Rongtao Xu, Zitong Huang, et~al.
\newblock Infiniteworld: A unified scalable simulation framework for general visual-language robot interaction.
\newblock \emph{arXiv preprint arXiv:2412.05789}, 2024.

\bibitem[Savva et~al.(2019)Savva, Kadian, Maksymets, Zhao, Wijmans, Jain, Straub, Liu, Koltun, Malik, et~al.]{savva2019habitat}
Manolis Savva, Abhishek Kadian, Oleksandr Maksymets, Yili Zhao, Erik Wijmans, Bhavana Jain, Julian Straub, Jia Liu, Vladlen Koltun, Jitendra Malik, et~al.
\newblock Habitat: A platform for embodied ai research.
\newblock In \emph{Proceedings of the IEEE/CVF international conference on computer vision}, pages 9339--9347, 2019.

\bibitem[Sferrazza et~al.(2024)Sferrazza, Huang, Lin, Lee, and Abbeel]{humanoidbench}
Carmelo Sferrazza, Dun-Ming Huang, Xingyu Lin, Youngwoon Lee, and Pieter Abbeel.
\newblock Humanoidbench: Simulated humanoid benchmark for whole-body locomotion and manipulation.
\newblock \emph{arXiv preprint arXiv:2403.10506}, 2024.

\bibitem[Sun et~al.(2025)Sun, Liu, Gu, Lim, Bhat, Tombari, Li, Haber, and Wu]{sun2024layoutvlm}
Fan-Yun Sun, Weiyu Liu, Siyi Gu, Dylan Lim, Goutam Bhat, Federico Tombari, Manling Li, Nick Haber, and Jiajun Wu.
\newblock Layoutvlm: Differentiable optimization of 3d layout via vision-language models.
\newblock \emph{CVPR}, 2025.

\bibitem[Sundaralingam et~al.(2023)Sundaralingam, Hari, Fishman, Garrett, Van~Wyk, Blukis, Millane, Oleynikova, Handa, Ramos, et~al.]{sundaralingam2023curobo}
Balakumar Sundaralingam, Siva Kumar~Sastry Hari, Adam Fishman, Caelan Garrett, Karl Van~Wyk, Valts Blukis, Alexander Millane, Helen Oleynikova, Ankur Handa, Fabio Ramos, et~al.
\newblock Curobo: Parallelized collision-free robot motion generation.
\newblock In \emph{2023 IEEE International Conference on Robotics and Automation (ICRA)}, pages 8112--8119. IEEE, 2023.

\bibitem[Tang et~al.(2024)Tang, Nie, Markhasin, Dai, Thies, and Nie{\ss}ner]{tang2023diffuscene}
Jiapeng Tang, Yinyu Nie, Lev Markhasin, Angela Dai, Justus Thies, and Matthias Nie{\ss}ner.
\newblock Diffuscene: Denoising diffusion models for gerative indoor scene synthesis.
\newblock In \emph{CVPR}, 2024.

\bibitem[Tao et~al.(2024)Tao, Xiang, Shukla, Qin, Hinrichsen, Yuan, Bao, Lin, Liu, Chan, et~al.]{tao2024maniskill3}
Stone Tao, Fanbo Xiang, Arth Shukla, Yuzhe Qin, Xander Hinrichsen, Xiaodi Yuan, Chen Bao, Xinsong Lin, Yulin Liu, Tse-kai Chan, et~al.
\newblock Maniskill3: Gpu parallelized robotics simulation and rendering for generalizable embodied ai.
\newblock \emph{arXiv preprint arXiv:2410.00425}, 2024.

\bibitem[Wang et~al.(2024)Wang, Chen, Huang, Ben, Wang, Mi, Huang, Zhao, Chen, Yang, et~al.]{wang2024grutopia}
Hanqing Wang, Jiahe Chen, Wensi Huang, Qingwei Ben, Tai Wang, Boyu Mi, Tao Huang, Siheng Zhao, Yilun Chen, Sizhe Yang, et~al.
\newblock Grutopia: Dream general robots in a city at scale.
\newblock \emph{arXiv preprint arXiv:2407.10943}, 2024.

\bibitem[Yang et~al.(2023)Yang, Zhang, Li, Zou, Li, and Gao]{yang2023som}
Jianwei Yang, Hao Zhang, Feng Li, Xueyan Zou, Chunyuan Li, and Jianfeng Gao.
\newblock Set-of-mark prompting unleashes extraordinary visual grounding in gpt-4v.
\newblock \emph{arXiv preprint arXiv:2310.11441}, 2023.

\bibitem[Yang et~al.(2024{\natexlab{a}})Yang, Jia, Zhi, and Huang]{yang2024physcene}
Yandan Yang, Baoxiong Jia, Peiyuan Zhi, and Siyuan Huang.
\newblock Physcene: Physically interactable 3d scene synthesis for embodied ai.
\newblock In \emph{CVPR}, 2024{\natexlab{a}}.

\bibitem[Yang et~al.(2024{\natexlab{b}})Yang, Sun, Weihs, VanderBilt, Herrasti, Han, Wu, Haber, Krishna, Liu, Callison-Burch, Yatskar, Kembhavi, and Clark]{Yang_2024_CVPR}
Yue Yang, Fan-Yun Sun, Luca Weihs, Eli VanderBilt, Alvaro Herrasti, Winson Han, Jiajun Wu, Nick Haber, Ranjay Krishna, Lingjie Liu, Chris Callison-Burch, Mark Yatskar, Aniruddha Kembhavi, and Christopher Clark.
\newblock Holodeck: Language guided generation of 3d embodied ai environments.
\newblock In \emph{CVPR}, 2024{\natexlab{b}}.

\bibitem[Yuan et~al.(2024)Yuan, Duan, Blukis, Pumacay, Krishna, Murali, Mousavian, and Fox]{robopoint}
Wentao Yuan, Jiafei Duan, Valts Blukis, Wilbert Pumacay, Ranjay Krishna, Adithyavairavan Murali, Arsalan Mousavian, and Dieter Fox.
\newblock Robopoint: A vision-language model for spatial affordance prediction for robotics.
\newblock \emph{arXiv preprint arXiv:2406.10721}, 2024.

\bibitem[Zhang et~al.(2025)Zhang, Yu, Lai, Lu, and Han]{zhang2025agentworld}
Yizheng Zhang, Zhenjun Yu, Jiaxin Lai, Cewu Lu, and Lei Han.
\newblock Agentworld: An interactive simulation platform for scene construction and mobile robotic manipulation.
\newblock \emph{arXiv preprint arXiv:2508.07770}, 2025.

\bibitem[Zheng et~al.(2022)Zheng, Chen, Jenkins, and Wang]{zheng2022vlmbench}
Kaizhi Zheng, Xiaotong Chen, Odest~Chadwicke Jenkins, and Xin Wang.
\newblock Vlmbench: A compositional benchmark for vision-and-language manipulation.
\newblock \emph{Advances in Neural Information Processing Systems}, 35:\penalty0 665--678, 2022.

\end{thebibliography}
}


\end{document}